\documentclass[conference]{IEEEtran}
\IEEEoverridecommandlockouts
\makeatletter
\def\thanks#1{\protected@xdef\@thanks{\@thanks
        \protect\footnotetext{#1}}}
\makeatother
\makeatletter
\renewcommand\footnoterule{\relax\kern-5pt
\hrule
\kern4.6pt}
\makeatother

\usepackage{amsmath, amssymb, graphicx}
\usepackage[nospace,noadjust]{cite}
\usepackage{multirow}
\usepackage{amsmath,amssymb,amsfonts}
\usepackage{algorithmic}
\usepackage{graphicx}
\usepackage{textcomp}
\usepackage{xcolor}
\usepackage{subcaption}
\def\BibTeX{{\rm B\kern-.05em{\sc i\kern-.025em b}\kern-.08em
    T\kern-.1667em\lower.7ex\hbox{E}\kern-.125emX}}

\usepackage{algorithm,algorithmic}

\def\bth{{\boldsymbol{\boldsymbol \theta}}}

\def\b1{{\boldsymbol{1}}}
\def\c1{{\textcircled{a}}}
\def\ba{{\boldsymbol{a}}}
\def\bb{{\boldsymbol{b}}}

\def\bn{{\boldsymbol{n}}}

\def\bu{{\boldsymbol{u}}}

\def\bx{{\boldsymbol{x}}}
\def\by{{\mathbf{y}}}
\def\bz{{\boldsymbol{z}}}
\def\bA{{\mathbf{A}}}

\def\bC{{\boldsymbol{C}}}

\def\bE{{\boldsymbol{E}}}
\def\bF{{\mathbf{F}}}

\def\bH{{\boldsymbol{H}}}
\def\bI{{\mathbf{I}}}

\def\bS{{\boldsymbol{S}}}
\def\bT{{\mathbf{T}}}

\newtheorem{theorem}{Theorem}[section]

\newtheorem{lemma}[theorem]{Lemma}

\begin{document}
%Copyright notice
%\pagestyle{empty}
\twocolumn[
\begin{@twocolumnfalse}

\large{\copyright  2025 IEEE.  Personal use of this material is permitted.  Permission from IEEE must be obtained for all other uses, in any current or future media, including reprinting/republishing this material for advertising or promotional purposes, creating new collective works, for resale or redistribution to servers or lists, or reuse of any copyrighted component of this work in other works.}\\
\\ %fill this for post-print

\end{@twocolumnfalse}
]
\title{MAP Image Recovery with Guarantees using Locally Convex Multi-Scale Energy (LC-MUSE) Model}

\author{Jyothi Rikhab Chand\textsuperscript{\textdagger,*}, Mathews Jacob\textsuperscript{*}\\
\textsuperscript{\textdagger}Department of Electrical and Computer Engineering, University of Iowa, IA, USA\\
\textsuperscript{*}Department of Electrical and Computer Engineering, University of Virginia, VA, USA}
\maketitle
\begin{abstract}
We propose a  multi-scale deep energy model that is strongly convex in the local neighbourhood around the data manifold to represent its probability density, with application in inverse problems. In particular, we represent the negative log-prior as a multi-scale energy model parameterized by a Convolutional Neural Network (CNN). We restrict the gradient of the CNN to be locally monotone, which constrains the model as a Locally Convex Multi-Scale Energy (LC-MuSE). We use the learned energy model in image-based inverse problems, where the formulation offers several desirable properties: i) uniqueness of the solution, ii) convergence guarantees to a minimum of the inverse problem, and iii) robustness to input perturbations. In the context of parallel Magnetic Resonance (MR) image reconstruction, we show that the proposed method performs better than the state-of-the-art convex regularizers, while the performance is comparable to plug-and-play regularizers and end-to-end trained methods.
\end{abstract}
\section{Introduction}\label{sec:intro}
\vspace{0.5mm}
Maximum A Posteriori (MAP) algorithms are widely used to recover images from noisy and undersampled measurements. These algorithms pose the reconstruction from undersampled data as a convex optimization problem, where the cost function is the sum of Data Consistency (DC) term and a hand-crafted prior, such as the Tikhonov regularizer \cite{tikhonov}. In many traditional MAP approaches, the regularizer is chosen to be strongly convex, which provides theoretical guarantees including uniqueness of the solution, robustness to input perturbations, and guaranteed convergence to a minimum of the cost function. These properties makes the convex MAP variants desirable in many practical applications.

Model-based Deep Learning (MoDL) methods, which offer improved performance and faster inference, were recently introduced. These methods are inspired by iterative Compressive Sensing (CS) algorithms, which alternate between a DC update and the proximal map of the regularizer. MoDL methods can be broadly grouped into two categories. The first category, known as the Plug-and-Play (PnP) method \cite{pnpbouman,consensus_equilibrium,rizwan_review},  replaces the proximal map used in CS algorithms with a pre-trained denoiser. PnP algorithms require the Convolutional Neural Network (CNN) based denoiser to be non-expansive to guarantee fixed-point convergence \cite{mol, pnp_ista}. Unfortunately, the enforcement of this constraint often restricts the achievable image quality, while not enforcing the constraint may result in divergence and visible artifacts in the reconstructed image \cite{muse}. The second category, unrolls the iterative steps of the CS algorithm and are trained in an End-to-End (E2E) fashion for a specific forward operator \cite{admmnet,variationalnet,modl,cinenet} and is found to have improved performance compared to the PnP method. However, the unrolling strategy is associated with significantly higher memory demand and computational complexity, making it difficult to apply to large-scale problems. In addition, the learned models are not easily transferable to other inverse problems, unlike the pre-trained PnP methods. 

The above MoDL approaches are not associated with an energy-based formulation. Energy-Based Models (EBMs), which explicitly model the negative log-prior as a CNN \cite{potential,gradientstep,muse}, were recently introduced for inverse problems. The main benefit of this scheme is the existence of a well-defined cost function, which enables the use of classical optimization methods that have guaranteed convergence, without the need for restrictive conditions on the CNN \cite{potential,gradientstep,muse}; the relaxation of the condition often translates to improved performance. Since EBMs model the prior distribution, this formulation can also be used in a Bayesian setting to sample from the posterior distribution and quantify the uncertainty estimate. This approach has been generalized to the implicit multi-scale setting in \cite{muse}. The use of multiple noise scales translated to improved convergence properties and consequently improved the performance, making the PnP energy framework comparable to E2E schemes.  

Despite the benefits, the above explicit energy models are not guaranteed to be convex; they cannot offer the above described theoretical guarantees that are enjoyed by CS algorithms. Several methods have been proposed to construct convex energy functions by imposing structural constraints on the network \cite{icnn,CRR}. For instance, the authors in \cite{icnn} proposed to learn a convex regularizer parameterized by the Input Convex Neural Network (ICNN), where the weights of some layers of the CNN are constrained to be non-negative with increasing and convex activation functions. Recently,  weakly convex regularizers \cite{wc1,wc2}, which also rely on architectural constraints have been proposed. The restriction on the network architecture translates to lower performance in image recovery tasks, when compared to non-convex EBM methods. To overcome these limitations, we propose to construct a local convex regularizer. In particular, we constrain the function to be strongly convex only in a local neighborhood of the data manifold. This is a significantly weaker constraint  and can be applied to arbitrary network architectures, unlike the restrictive ICNN framework. Consequently, we expect the relaxation to translate to improved performance. We note that the local nature of the convexity constraint is not a limitation in practical applications. Typically, image reconstruction algorithms are initialized using reasonable guesses that are fast to compute. Therefore, when the initializations are within a local neighborhood, we show that the corresponding algorithms offer guaranteed uniqueness, robustness, and convergence. We leverage the fact that a function is strictly convex in a domain iff its derivative is monotone within that domain. We encourage the local monotone constraint using an adversarial regularization strategy. We compare the performance of the LC-MuSE approach with state-of-the-art methods in the context of MR image recovery from undersampled measurements. Despite the focus on the MR imaging application, the proposed framework is generally applicable to general computational imaging problems. 
\section{Background}
We consider the recovery of unknown image $\bx \in \mathbb{C}^m$ from its noisy and undersampled measurements $\bb \in \mathbb{C}^n$  such that $\bb = \bA\bx +\bn$, where $\bn \sim \mathcal{N} (0, \eta\bI)$ and $\bA \in \mathbb{C}^{n \times m}$ is a linear forward operator. Then the MAP estimate of $\bx$ is given as:  
\begin{equation}\label{eq:P1}
  \bx^*(\bb)
=\arg \min_{\bx} f(\bx,\bb)= \underbrace{\dfrac{1}{2\eta^{2}} \|\bA\bx-\bb\|_{2}^2}_{-\log p(\bb|\bx)} -\log p(\bx) 
\end{equation}
where $p(\bb|\bx)$ and $p(\bx)$ denote the likelihood of the measurement and the prior distribution of the image $\bx$, respectively. In this work, we use the energy $\bE_\theta (\bx): \mathbb{C}^m \rightarrow \mathbb{R}^{+}$ to model the negative log-prior:
\begin{eqnarray}\label{e1}
    \bE_\bth(\bx) &=& \dfrac{1}{2\sigma_f^2}{\|\bx-{\Psi}_{\bth}(\bx)\|^{2}}
\end{eqnarray}
Here $\Psi_\bth: \mathbb C^m\rightarrow \mathbb C^m$ is CNN network, $\sigma_f >0 $ is a parameter used in training the network. The score function of the energy is essentially the gradient: $ \bH_\bth(\bx)=\nabla_\bx \boldsymbol E_\bth(\bx)$.% is specified as: 
%\begin{eqnarray}\label{score}
%	\bH_\bth(\bx)  &=& \left(\bx - \Psi_{\bth}\left(\bx\right)\right)- \nabla_{\bx}\Psi_{\bth}\left(\bx- \Psi_{\bth}(\bx)\right)
%\end{eqnarray}
The energy function $\bE_\bth(\cdot)$ can be pre-trained using the multiscale score matching approach \cite{muse,vincent2010}:
\begin{eqnarray}\label{DSM}
\bth^* 
%&=& 
= \arg \min_{\bth} \underbrace{\mathbb E_{\sigma}\left(\mathbb E_{\bx} \mathbb E_{\bz}  \left\|\nabla_\bx {\bE}_{\theta} (\bx+\sigma \bz) - \sigma \bz  \right\|_{2}^2\right)}_{J_\bth(\bx)}
\end{eqnarray}
where $\bz \sim \mathcal{N}(0,\bI)$ and $\sigma$ is noise standard deviation which is chosen randomly. We note that \eqref{DSM} encourages the score to learn the added noise $\sigma \bz$. 

The multi-scale training strategy translates into iterative MAP algorithms with convergence to a minimum of \eqref{eq:P1}, providing image quality comparable to the methods trained in an E2E fashion \cite{muse}. However, the energy function trained using the score matching objective function in \eqref{DSM} is not guaranteed to be convex. We note that the convexity of the energy function will ensure guaranteed uniqueness and robustness of the solution, as well as convergence of the algorithm. Hence, we focus on constructing a convex multi-scale energy function with these goals. As discussed earlier, the use of a constrained CNN architecture similar to \cite{icnn} to ensure convexity often translates into reduced performance. An alternative is to use the following properties to ensure the convexity of the energy function.
\begin{lemma}\label{lemma1}
$\bE_\bth(\cdot)$ is m-strongly convex iff  $\bH_\bth (\bx)=\nabla_\bx \bE_\bth (\bx)$ is m-strongly monotone \cite{monotone_prop}:
\begin{eqnarray}\nonumber
\left\langle\bH_\bth(\bx)-\bH_\bth(\by),\bx-\by)\right\rangle &\geq& m\|\bx-\by\|^{2};m>0; \\&& \forall \bx, \by \in \mathbb{C}^m
\end{eqnarray}
\end{lemma}
Here, the dot product $\left\langle\ba,\bb\right\rangle = \rm{Re}\left(\ba^H\bb\right)$. The following result can be used to constrain the score function $\bH_\bth$ as an m-monotone operator.
\begin{lemma}\label{lemma2}
The score function $\bH_\bth$ is m-strongly monotone if the Lipschitz constant of  $\bT_\bth= \bI-\bH_\bth$ is $1-m$ \cite{mol}.
\end{lemma}
While one may use spectral normalization technique to impose the above constraint \cite{sn}, it imposes restrictions on the architectures that can be used as well as the network weights, again translating to reduced performance. 

\section{Proposed method}

The main focus of this work is to derive a MAP algorithm with good performance as well as guaranteed uniqueness, robustness, and convergence properties. We relax the global convexity assumed in Lemma \ref{lemma1} to local convexity around the image manifold. We define a function $\bE_\theta(\cdot)$ to be m-strong Locally Convex (LC) in a ball of radius $\delta$, centered at the point $\bu$:
\begin{equation}
 \mathcal{B}_\delta(\bu)  = \{\tilde{\bu}: \|\tilde{\bu}-\bu\| \leq \delta \},
\end{equation}
if it satisfies:
\begin{eqnarray}\label{local_convx_def}\nonumber
    \bE_\bth(\bx) &\geq& \bE_\bth(\by) + \left\langle\bH_\bth(\by),(\bx - \by)\right\rangle + \dfrac{m}{2}\|\bx -\by\|^{2}  \\&& \:\:\qquad \forall \bx,\by \in \mathcal{B}_\delta(\bu), m>0, 
\end{eqnarray}
%We illustrate the concept of a local convex function in Fig.\ref{fig_slc}.
We note that as $\delta \rightarrow \infty$, $E_\bth(\bx)$ is globally convex. The use of the weaker m-strong LC constraint on the energy model is expected to translate to improved performance compared to the ICNN approach, which requires specific neural network architecture. 

We now rely on the extensions of Lemmas \ref{lemma1}-\ref{lemma2} to the LC setting to constrain the energy as a m-strong LC function. 
%Note that if $m=0$, the above function would be LC around $\bu$ with radius $\delta$.
% \begin{figure}
%     \centering
% \includegraphics[width=0.9\linewidth]{results/local_convex_illustration.pdf}
%     \caption{Local Convex function illustration where we restrict the function to be convex in a ball of radius $\delta$ centered at $\bx^{*}$. \textcolor{red}{Is this figure needed ?? Remove this to save space ?}}
%     \label{fig_slc}
% \end{figure}
\begin{lemma}
$\bE_\bth(\cdot)$ is m-strong LC in $\mathcal{B}_\delta (\bu)$ iff its gradient is m-strong Locally Monotone (LM) in $\mathcal{B}_\delta (\bu)$ defined as: 
\begin{eqnarray}\label{local_monotone_def}
\left\langle\bH_\bth(\bx)-\bH_\bth(\by),\bx-\by\right\rangle\geq m\|\bx-\by\|^{2}; \forall \bx, \by \in \mathcal{B}_\delta (\bu).
\end{eqnarray}
\end{lemma}
We omit the proof because of space constraints. 
\begin{lemma}
The score function $\bH_\bth$ is m-strong LM in $\mathcal{B}_\delta(\bu)$ if the local Lipschitz constant of $\bT_\bth =\bI-\bH_\bth$ satisfies:
\begin{eqnarray}
    \max_{\bx,\by \in \mathcal{B}_\delta(\bu)} \dfrac{\|\bT_\bth(\bx) -\bT_\bth(\by)\|}{\|\bx-\by\|} \leq l= (1-m)
\end{eqnarray}
\end{lemma}
\subsection{Training Procedure}
 
Lemmas 3.1 and 3.2 show that if we can constrain the local Lipschitz constant of $\bT_\bth$, the corresponding energy is m-strong LC. We introduce a training procedure to learn the energy model that satisfies this constraint:
%\begin{eqnarray}\label{DSM_constraint}
% \bth^*
%=\arg \min_{\bth}   J_\bth (\bx) \quad \textrm{s.t.} \:\: L[\bT_\bth(\bx)] \leq L_T
% \end{eqnarray}
 \begin{equation}\label{DSM_constraint}
   \bth^*
=\arg \min_{\bth}   J(\bth) +\lambda ~\mathbb E_{\bx}\Big(R( L[\bT_\bth(\bx)]- l)\Big)^2,
\end{equation}
where $J_\bth(\bx)$ is defined in \eqref{DSM} and $R(\bx)$ is the Rectified Linear Unit (ReLU) activation applied on $\bx$. We note that as $\lambda\rightarrow \infty$, the constraint $L[\bT_\bth(\bx)] \leq l$ will be enforced exactly. Following \cite{CLIP,mol}, we estimate the local Lipschitz constant $L[\bT_\bth(\bx)]$ numerically within the training loop as:
 \begin{eqnarray}\label{local_lipschitz}
     L[\bT_\bth(\bx)]=\max_{\bx_1,\bx_2 \in \mathcal{B}_\delta(\bx)} \dfrac{\|\bT_\bth(\bx_1)-\bT_\bth(\bx_2))\|}{\|\bx_1-\bx_2\|}
 \end{eqnarray}
where $\mathcal{B}_\delta(\bx)$ denotes a ball of radius $\delta$ centered at the training data point $\bx$. At every iteration, we estimate $L[\bT_\bth(\bx)]$ (for a fixed $\bth$) by solving \eqref{local_lipschitz}  using projected gradient ascent algorithm. This algorithm is initialized randomly such that $\bx_1$ and $\bx_2$ are within $\mathcal{B}_\delta(\bx)$. At every iteration, $\bx_1$ and $\bx_2$ is updated using gradient ascent step and projected to $\mathcal{B}_\delta(\bx)$ in \eqref{local_lipschitz}.  We refer the proposed regularizer trained using \eqref{DSM_constraint} as Locally Convex Multi-Scale Energy (LC-MuSE) model.
\subsection{MAP estimation}
Once the energy function is learned, we use it in the MAP objective \eqref{eq:P1} iteratively using the Majorization Minimization (MM) framework.  Following the steps in \cite{muse}, we use the following update rule to get the MAP estimate:
\begin{equation}\label{mm_imuse}
        \bx_{n+1} = \left(\dfrac{\bA^{H}\bA}{\zeta^{2}}+L~\bI\right)^{-1}\left(\dfrac{\bA^{H}\bb}{\zeta^{2}} + L \bx_{n} - \nabla_{\bx_{n} }\bE_{\bth}(\bx_{n})\right)
\end{equation}  
where $\zeta= \eta/\sigma_f$, $\sigma_f$ is the finest noise standard deviation level used to train $\bE_\theta(\cdot)$, and L is its Lipschitz constant.

\subsection{Theoretical Guarantees}
We note that the $\bA$ operator in \eqref{eq:P1} is linear and hence $\bA^H\bA \succeq 0$. This implies that $f(\bx,\bb)$ in \eqref{eq:P1} which is the  sum of a convex log-likelihood term and m-strong LC function $\bE_\bth(\cdot)$, is a m-strong LC function. We now show that the m-strong LC energy formulation translates to the following desirable theoretical properties.

\begin{lemma}[Uniqueness]
The solution of \eqref{eq:P1} is unique within $\mathcal{B}_\delta(\bx)$. 
\end{lemma}

\begin{lemma}\label{conv}
    If the MM-based algorithm is initialized with an $\bx_0 \in \mathcal{B}_\delta(\bx^*)$, then the sequence of iterates will converge to a unique minimum of \eqref{eq:P1}.
\end{lemma}
%\begin{IEEEproof}
%Refer \cite{mm_conv} for the proof of convergence of \eqref{mm_imuse} to a stationary point. Note that since  \eqref{eq:P1} is m-strong LC, this would translate to the algorithm  converging to a global minimum provided $\bx_0 \in \mathcal{B}_\delta(\bx^*)$. 
%\end{IEEEproof}

\begin{lemma}
   Let $\bx^{*}(\bb)$ and  $\bx^{*}(\bb+\bn)$ denote the minimizers corresponding to the measurements $\bb$ and the perturbed measurements $\bb + \bn$. Then the following holds:
   \begin{equation}
       \|\bx^*(\bb+\bn)-\bx^*(\bb)\| \leq \dfrac{\|\bn\|  }{m \eta^2},
   \end{equation}
   provided $\|\bn\| \leq {m} \delta \eta^2$.
\end{lemma}
We note that the norm of the upperbound is inversely proportional to $m$. Therefore, a higher value of $m$ will give a more robust algorithm. 
\section{Results}
\subsection{Data set}
We demonstrate the performance of LC-MuSE in the context of 2D MR image construction. In this case, the linear operator in \eqref{eq:P1} is defined as $\bA=\bS\bF\bC$, where $\bS$ is the sampling matrix, $\bF$ is the Fourier matrix, and $\bC$ is the Coil Sensivitiy Map (CSM) that is estimated using the algorithm in \cite{espirit}. The MR images were obtained from the publicly available parallel fastMRI T2-weighted brain data set \cite{knee_dataset}. It consists of a twelve-channel brain data with  complex images of size $320 \times 320 $. The data set was split into $45$ training, $5$ validation, and $50$ test subjects. We evaluate the reconstruction algorithms for different acceleration factors using 1D and 2D Cartesian undersampling masks. 
\subsection{Architecture and implementation}
We use a five-layer CNN to realize ${\Psi}_{\bth}(\bx)$ in \eqref{e1}. In particular, each layer consists of $64$ channels with $3\times 3$ filter size. We used ReLU between each layer, except for the final layer. Both MuSE \cite{muse} and LC-MuSE followed the above architecture.  Nonetheless, MuSE was trained using \eqref{DSM} while LC-MuSE was trained using \eqref{DSM_constraint}, where the standard deviations are chosen randomly from uniform distribution ranging from $0$ to $0.1$. We propose to initialize LC-MuSE with the SENSE reconstruction \cite{sense}: $\bx_0 = (\bA^H\bA + \tilde{\lambda}\bI)^{-1}(\bA^H \bb)$. We note that according to lemma \ref{conv}, the MM algorithm in \eqref{mm_imuse} is guaranteed to converge to a global minimum if $\|\bx_0 -\bx^*\| \leq \delta$. Therefore, we choose $\delta$ as the worst-case deviation of the SENSE solutions from the reference images, for all the images in the training data set. 

\begin{figure*}[!htp]
        \centering
        \begin{subfigure}[b]{0.58\textwidth}
            \centering
            \includegraphics[width=\textwidth]{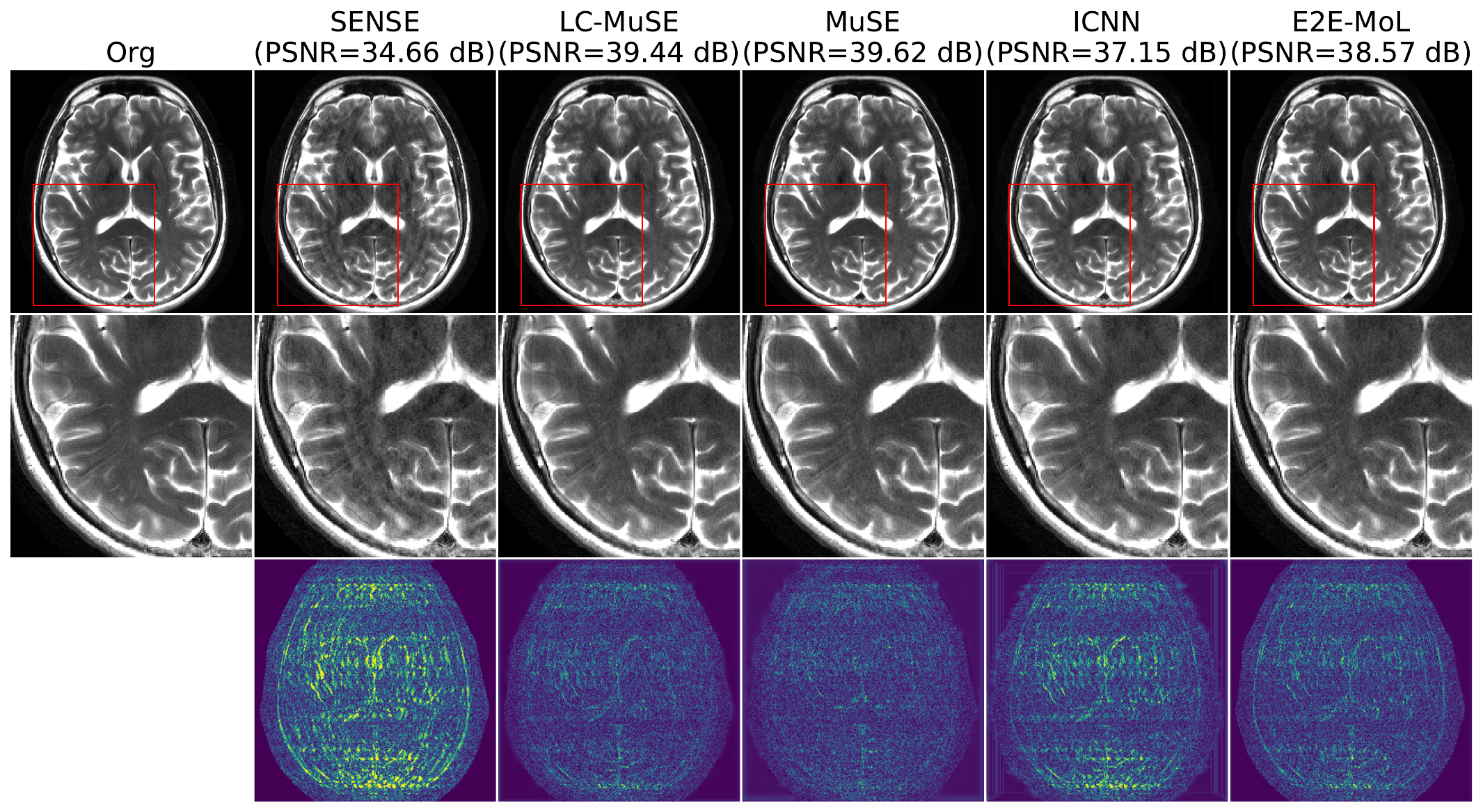}  
            \caption{}
            \label{two_fold}
        \end{subfigure}
       % \begin{subfigure}[b]{0.475\textwidth}
       %      \centering
       %      \includegraphics[width=\textwidth]{results/conv_1D_two_fold_sense.pdf}
       %      \caption{1D mask and two-fold acceleration}  
       %      \label{two_fold}
       %  \end{subfigure}
        \hfill
        \begin{subfigure}[b]{0.58\textwidth}  
            \centering 
            \includegraphics[width=\textwidth]{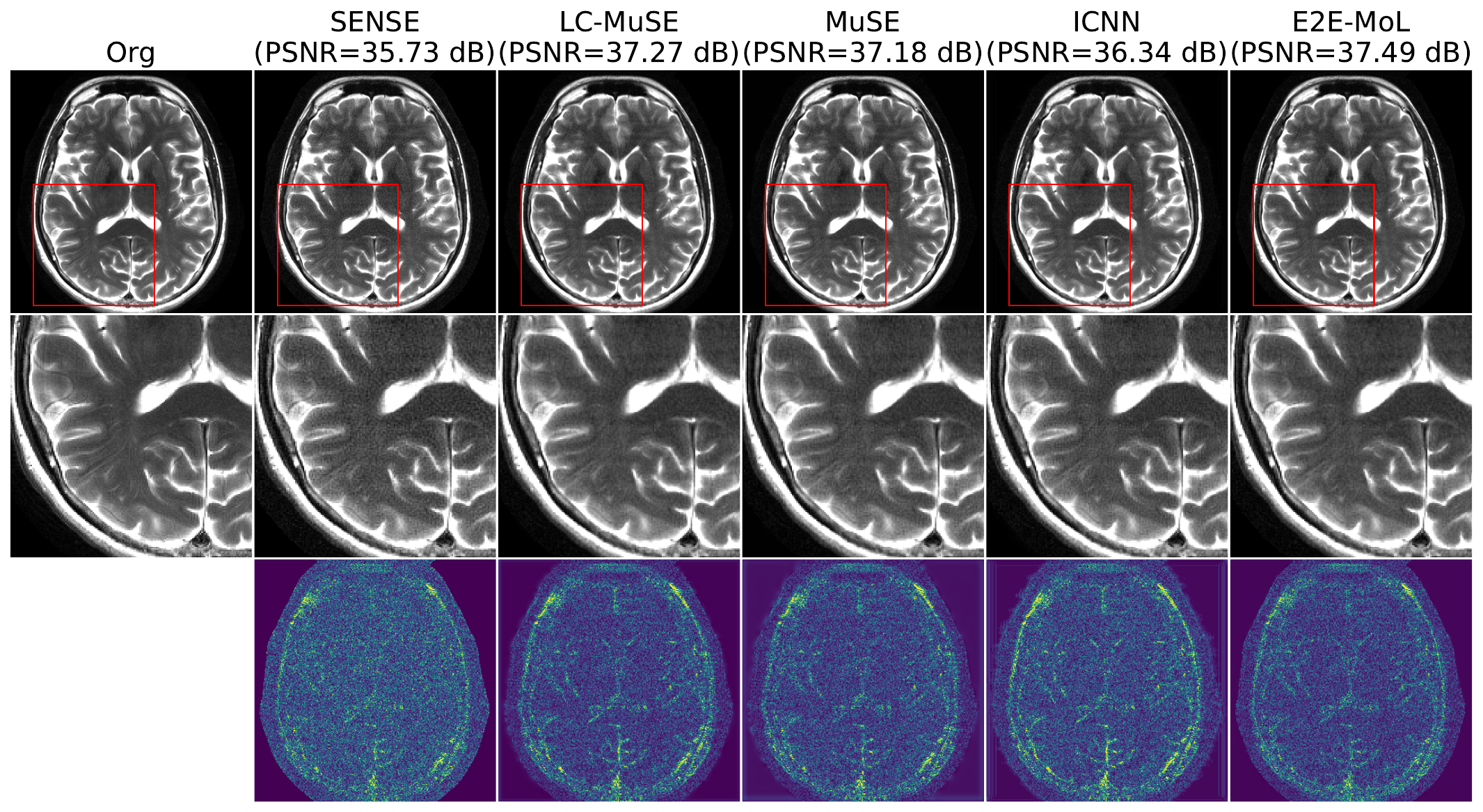} 
            \caption{}
            \label{six_fold}
        \end{subfigure}
        % \begin{subfigure}[b]{0.475\textwidth}
        %     \centering
        %     \includegraphics[width=\textwidth]{results/conv_2D_six_fold_sense.pdf}
        %     \caption{2D mask and six-fold acceleration}  
        %     \label{two_fold}
        % \end{subfigure}
        \caption{Comparison of LC-MuSE with MuSE, ICNN and E2E-trained MoL at a) two-fold and b) six-fold acceleration using 1D and 2D undersampling mask, respectively. The top, middle, and last rows show the reconstructed, enlarged, and error images, respectively. }
        %\textcolor{red}{Do you have a message on the #iterations ?? If not, better to not include this. Convergence in one step may raise some suspicion. }} 
        \label{recon}
\end{figure*}
\subsection{State-of-the-art (SOTA) methods for comparison}
We compared the proposed approach to ICNN \cite{icnn}, where $\bE_\bth(\cdot)$ is constrained to be globally convex by a specific choice of architecture  \cite{icnn}. %To do so we construct $\bE_\bth(\cdot)$ as ICNN \cite{icnn} which is built using the fact that: a) non-negative combination of convex function is another convex function and, b) the composition of two functions $\phi_1$ and $\phi_2$ is convex, if $\phi_2$ is convex and $\phi_1$ is convex and non-decreasing. As one may observe, this network is very restrictive compared to ${\Psi}_{\bth}(\bx)$. 
For fairness, we kept the number of filters per layer and number of layers similar to the proposed model. %Out of the seven CNN layers, we restricted three of them to have non-negative weights. The activation function between each layer (except the final layer) was chosen as Leaky-ReLU with a negative slope of $0.2$. At the final layer, linear activation function was used. ICNN was also trained using \eqref{DSM} with  standard deviations chosen randomly from uniform distribution ranging from $0$ to $0.1$. 
We also compared the above pre-trained algorithms with MoL \cite{mol} which was trained in an E2E-fashion using the deep equilibrium framework \cite{deq}. We used a five-layer CNN  to implement MoL and the log-barrier approach was used to constraint the Lipschitz constant as in \cite{mol}. All the algorithms were initialized with SENSE.
\subsection{Experimental results}
%In this section, we compare the reconstruction performance of the algorithms on the test data for two-fold and four-fold accelerations which was undersampled using 1D and 2D mask, respectively. All the algorithms were initialized with SENSE. Except for E2E-MoL, the remaining methods implemented the MM algorithm with update step \eqref{mm_imuse} and the algorithm was run until $\dfrac{|f(\bx_{n+1})-f(\bx_n)}{f(\bx_n)} \leq 1e^{-6}$ or until $500$ iterations were reached. The optimal $\zeta$ for the algorithms were chosen based on their performance on the validation set. 

Table \ref{recon_cmp} compares the reconstruction performance for two different acquisition settings: 2x and 6x acceleration using 1D and 2D undersampling masks. PSNR and SSIM are used as evaluation metrics. Figs. \ref{recon}a and \ref{recon}b show the reconstructed images for two-fold and four-fold accelerations, respectively. We observe that for both the acquisition settings, LC-MuSE offers better performance than ICNN, which is a more constrained model. We also note that the performance of ICNN is comparable to that of MuSE and the E2E optimized MoL approach. These results show that the proposed framework offers guaranteed image recovery similar to that of compressive sensing, while the performance is comparable to SOTA deep learning methods. 
%On the other hand, the improved performance of LC-MuSE stems from the fact that it only requires its score to be locally monotone which is a much weaker restriction on the weights of the network.  
\begin{table}[htp]
\caption{Quantitative comparison of reconstruction performance on the fastMRI brain dataset for two different accelerations.}
\centering
\begin{tabular}{|l|p{0.1cm}|p{0.5cm}|p{0.5cm}|p{0.5cm}|}
\hline
\multirow{2}{*}{Algorithm} & \multicolumn{2}{c|}{2x with 1D mask} & \multicolumn{2}{c|}{6x with 2D mask} \\ \cline{2-5}
                           & \multicolumn{1}{l|}{PSNR (dB)} & SSIM & \multicolumn{1}{l|}{PSNR (dB)} & SSIM \\ \hline

SENSE & \multicolumn{1}{l|}{$34.29 \pm 1.34$} &0.95  & \multicolumn{1}{l|}{$35.08 \pm 1.23$} &$0.95$  \\ 
LC-MuSE     & \multicolumn{1}{l|}{$38.34 \pm 1.96$} & 0.97& \multicolumn{1}{l|}{$36.98 \pm 1.27$} & 0.97 \\ 
ICNN                       & \multicolumn{1}{l|}{$36.49 \pm 1.86$} & 0.95& \multicolumn{1}{l|}{$35.97 \pm 1.15$} &$0.95$  \\ 
MuSE         & \multicolumn{1}{l|}{$38.39 \pm 1.91$} & 0.97 & \multicolumn{1}{l|}{$36.61 \pm 1.19$} &$0.96$  \\ 
E2E-MoL                    & \multicolumn{1}{l|}{$37.80 \pm 1.70$} & 0.97 & \multicolumn{1}{l|}{$36.91 \pm 1.19$} &0.97  \\ \hline
\end{tabular}\label{recon_cmp}
\end{table}
\section{Conclusion}
We proposed a locally convex multi-scale energy regularizer, which does not require any explicit structural constraint on the network. We showed that the proposed regularizer, under mild conditions, has several desirable properties. The reconstruction results shows that the proposed methods perform better than the existing convex regularizer and has comparable performance with non-convex PnP regularizer and E2E-trained method.
\section{Compliance with ethical standards}
\label{sec:ethics}

This study was conducted on a publicly available human subject data set. Ethical approval was not required, as confirmed by the license attached with the open-access data.

\section{Acknowledgments}
\label{sec:acknowledgments}

This work is supported by NIH grants R01-AG067078, R01-EB031169, and R01-EB019961. The funding organization had no role in the design, conduct, analysis, or publication of this research.
\bibliographystyle{IEEEtran}
\bibliography{refs}

% Generated by IEEEtran.bst, version: 1.14 (2015/08/26)
\begin{thebibliography}{10}
\providecommand{\url}[1]{#1}
\csname url@samestyle\endcsname
\providecommand{\newblock}{\relax}
\providecommand{\bibinfo}[2]{#2}
\providecommand{\BIBentrySTDinterwordspacing}{\spaceskip=0pt\relax}
\providecommand{\BIBentryALTinterwordstretchfactor}{4}
\providecommand{\BIBentryALTinterwordspacing}{\spaceskip=\fontdimen2\font plus
\BIBentryALTinterwordstretchfactor\fontdimen3\font minus
  \fontdimen4\font\relax}
\providecommand{\BIBforeignlanguage}[2]{{%
\expandafter\ifx\csname l@#1\endcsname\relax
\typeout{** WARNING: IEEEtran.bst: No hyphenation pattern has been}%
\typeout{** loaded for the language `#1'. Using the pattern for}%
\typeout{** the default language instead.}%
\else
\language=\csname l@#1\endcsname
\fi
#2}}
\providecommand{\BIBdecl}{\relax}
\BIBdecl

\bibitem{tikhonov}
A.~N. Tikhonov and V.~I. Arsenin, \emph{Solutions of Ill-posed Problems: Andrey
  N. Tikhonov and Vasiliy Y. Arsenin. Translation Editor Fritz John}.\hskip 1em
  plus 0.5em minus 0.4em\relax Wiley, 1977.

\bibitem{pnpbouman}
S.~V. Venkatakrishnan, C.~A. Bouman, and B.~Wohlberg, ``Plug-and-play priors
  for model based reconstruction,'' in \emph{2013 IEEE Global Conference on
  Signal and Information Processing}, 2013, pp. 945--948.

\bibitem{consensus_equilibrium}
G.~T. Buzzard, S.~H. Chan, S.~Sreehari, and C.~A. Bouman, ``Plug-and-play
  unplugged: Optimization-free reconstruction using consensus equilibrium,''
  \emph{SIAM Journal on Imaging Sciences}, vol.~11, no.~3, pp. 2001--2020,
  2018.

\bibitem{rizwan_review}
R.~Ahmad, C.~A. Bouman, G.~T. Buzzard, S.~Chan, S.~Liu, E.~T. Reehorst, and
  P.~Schniter, ``Plug-and-play methods for magnetic resonance imaging: Using
  denoisers for image recovery,'' \emph{IEEE signal processing magazine},
  vol.~37, no.~1, pp. 105--116, 2020.

\bibitem{mol}
A.~Pramanik, M.~B. Zimmerman, and M.~Jacob, ``Memory-efficient model-based deep
  learning with convergence and robustness guarantees,'' \emph{IEEE
  Transactions on Computational Imaging}, vol.~9, pp. 260--275, 2023.

\bibitem{pnp_ista}
E.~Ryu, J.~Liu, S.~Wang, X.~Chen, Z.~Wang, and W.~Yin, ``Plug-and-play methods
  provably converge with properly trained denoisers,'' in \emph{International
  Conference on Machine Learning}.\hskip 1em plus 0.5em minus 0.4em\relax PMLR,
  2019, pp. 5546--5557.

\bibitem{muse}
J.~R. Chand and M.~Jacob, ``Multi-scale energy (muse) framework for inverse
  problems in imaging,'' \emph{IEEE Transactions on Computational Imaging}, pp.
  1--16, 2024.

\bibitem{admmnet}
J.~Sun, H.~Li, Z.~Xu \emph{et~al.}, ``Deep admm-net for compressive sensing
  mri,'' \emph{Advances in neural information processing systems}, vol.~29,
  2016.

\bibitem{variationalnet}
K.~Hammernik, T.~Klatzer, E.~Kobler, M.~P. Recht, D.~K. Sodickson, T.~Pock, and
  F.~Knoll, ``Learning a variational network for reconstruction of accelerated
  mri data,'' \emph{Magnetic resonance in medicine}, vol.~79, no.~6, pp.
  3055--3071, 2018.

\bibitem{modl}
H.~K. Aggarwal, M.~P. Mani, and M.~Jacob, ``Modl: Model-based deep learning
  architecture for inverse problems,'' \emph{IEEE transactions on medical
  imaging}, vol.~38, no.~2, pp. 394--405, 2018.

\bibitem{cinenet}
T.~K{\"u}stner, N.~Fuin, K.~Hammernik, A.~Bustin, H.~Qi, R.~Hajhosseiny, P.~G.
  Masci, R.~Neji, D.~Rueckert, R.~M. Botnar \emph{et~al.}, ``Cinenet: deep
  learning-based 3d cardiac cine mri reconstruction with multi-coil
  complex-valued 4d spatio-temporal convolutions,'' \emph{Scientific reports},
  vol.~10, no.~1, p. 13710, 2020.

\bibitem{potential}
R.~Cohen, Y.~Blau, D.~Freedman, and E.~Rivlin, ``It has potential:
  Gradient-driven denoisers for convergent solutions to inverse problems,''
  \emph{Advances in Neural Information Processing Systems}, vol.~34, pp.
  18\,152--18\,164, 2021.

\bibitem{gradientstep}
S.~Hurault, A.~Leclaire, and N.~Papadakis, ``Gradient step denoiser for
  convergent plug-and-play,'' \emph{arXiv preprint arXiv:2110.03220}, 2021.

\bibitem{icnn}
S.~Mukherjee, S.~Dittmer, Z.~Shumaylov, S.~Lunz, O.~{\"O}ktem, and C.-B.
  Sch{\"o}nlieb, ``Learned convex regularizers for inverse problems,''
  \emph{arXiv preprint arXiv:2008.02839}, 2020.

\bibitem{CRR}
A.~Goujon, S.~Neumayer, P.~Bohra, S.~Ducotterd, and M.~Unser, ``A
  neural-network-based convex regularizer for inverse problems,'' \emph{IEEE
  Transactions on Computational Imaging}, 2023.

\bibitem{wc1}
A.~Goujon, S.~Neumayer, and M.~Unser, ``Learning weakly convex regularizers for
  convergent image-reconstruction algorithms,'' \emph{SIAM Journal on Imaging
  Sciences}, vol.~17, no.~1, pp. 91--115, 2024.

\bibitem{wc2}
Z.~Shumaylov, J.~Budd, S.~Mukherjee, and C.-B. Sch{\"o}nlieb, ``Weakly convex
  regularisers for inverse problems: Convergence of critical points and
  primal-dual optimisation,'' \emph{arXiv preprint arXiv:2402.01052}, 2024.

\bibitem{vincent2010}
P.~Vincent, ``A connection between score matching and denoising autoencoders,''
  \emph{Neural computation}, vol.~23, no.~7, pp. 1661--1674, 2011.

\bibitem{monotone_prop}
E.~K. Ryu and W.~Yin, \emph{Large-scale convex optimization: algorithms \&
  analyses via monotone operators}.\hskip 1em plus 0.5em minus 0.4em\relax
  Cambridge University Press, 2022.

\bibitem{sn}
T.~Miyato, T.~Kataoka, M.~Koyama, and Y.~Yoshida, ``Spectral normalization for
  generative adversarial networks,'' \emph{arXiv preprint arXiv:1802.05957},
  2018.

\bibitem{CLIP}
L.~Bungert, R.~Raab, T.~Roith, L.~Schwinn, and D.~Tenbrinck, ``Clip: Cheap
  lipschitz training of neural networks,'' in \emph{International Conference on
  Scale Space and Variational Methods in Computer Vision}.\hskip 1em plus 0.5em
  minus 0.4em\relax Springer, 2021, pp. 307--319.

\bibitem{espirit}
M.~Uecker, P.~Lai, M.~J. Murphy, P.~Virtue, M.~Elad, J.~M. Pauly, S.~S.
  Vasanawala, and M.~Lustig, ``Espirit—an eigenvalue approach to
  autocalibrating parallel mri: where sense meets grappa,'' \emph{Magnetic
  resonance in medicine}, vol.~71, no.~3, pp. 990--1001, 2014.

\bibitem{knee_dataset}
J.~Zbontar, F.~Knoll, A.~Sriram, T.~Murrell, Z.~Huang, M.~J. Muckley,
  A.~Defazio, R.~Stern, P.~Johnson, M.~Bruno \emph{et~al.}, ``fastmri: An open
  dataset and benchmarks for accelerated mri,'' \emph{arXiv preprint
  arXiv:1811.08839}, 2018.

\bibitem{sense}
K.~P. Pruessmann, M.~Weiger, M.~B. Scheidegger, and P.~Boesiger, ``Sense:
  sensitivity encoding for fast mri,'' \emph{Magnetic Resonance in Medicine: An
  Official Journal of the International Society for Magnetic Resonance in
  Medicine}, vol.~42, no.~5, pp. 952--962, 1999.

\bibitem{deq}
S.~Bai, J.~Z. Kolter, and V.~Koltun, ``Deep equilibrium models,''
  \emph{Advances in Neural Information Processing Systems}, vol.~32, 2019.

\end{thebibliography}

\end{document}